\documentclass[letterpaper, 10 pt, conference]{ieeeconf}  
\usepackage{amsmath,amsfonts}
\usepackage{algorithm}
\usepackage{algorithmicx}
\usepackage{algpseudocode}
\usepackage{array}
\usepackage[caption=false,font=footnotesize,labelfont=rm,textfont=rm]{subfig}
\usepackage{textcomp}
\usepackage{stfloats}
\usepackage{url}
\usepackage{verbatim}
\usepackage{graphicx}
\usepackage{multirow}
\usepackage{booktabs}
\usepackage{makecell}
\usepackage[noadjust]{cite}

\renewcommand{\a}{\mathbf{a}}

\newcommand{\e}{\mathbf{e}}

\newcommand{\g}{\mathbf{g}}

\newcommand{\p}{\mathbf{p}}
\newcommand{\R}{\mathbf{R}}

\newcommand{\s}{\mathbf{s}}

\newcommand{\vel}{\mathbf{v}}

\newcommand{\bk}{\mathbf{K}}

\newcommand{\bomega}{\boldsymbol{\omega}}
\newcommand{\btau}{\boldsymbol{\tau}}

\graphicspath{{figures/}}

\IEEEoverridecommandlockouts

\title{\LARGE \bf\textbf{
TACO: General Acrobatic Flight Control via Target-and-Command-Oriented Reinforcement Learning 
}}
\author{Zikang Yin$^{1,2}$, Canlun Zheng$^{1,2}$, Shiliang Guo$^2$, Zhikun Wang$^{2,*}$, Shiyu Zhao$^2$ 
\thanks{$^1$College of Computer Science and Technology, Zhejiang University, Hangzhou, China. $^2$WINDY Lab, Department of Artificial Intelligence, Westlake University, Hangzhou, China. \{yinzikang, zhengcanlun, guoshiliang, wangzhikun, zhaoshiyu\}@westlake.edu.cn. *Corresponding author.}
}

\begin{document}
\maketitle

\begin{abstract}
Although acrobatic flight control has been studied extensively, one key limitation of the existing methods is that they are usually restricted to specific maneuver tasks and cannot change flight pattern parameters online. In this work, we propose a target-and-command-oriented reinforcement learning (TACO) framework, which can handle different maneuver tasks in a unified way and allows online parameter changes. We also propose a spectral normalization method with input-output rescaling to enhance the policy's temporal and spatial smoothness, independence, and symmetry, thereby overcoming the sim-to-real gap. We validate the TACO approach through extensive simulation and real-world experiments, demonstrating its ability to achieve high-speed, high-accuracy circular flights and continuous multi-flips. 

\end{abstract}


\section{Introduction}

This paper studies the task of aerobatic flight control of MAVs (micro aerial vehicles). 
While this task has many important applications in practice~\cite{liAggressiveOnlineControl2020,chenAerialGraspingVelocity2022}, our research is specifically motivated by agile MAV-Capture-MAV \cite{yuCatchPlannerCatching2023, liThreeDimensionalBearingOnlyTarget2022}, where one or multiple MAVs detect, localize, follow, and eventually capture a target MAV.
We have proposed cooperative MAV-Capture-MAV systems~\cite{zhengOptimalSpatialTemporalTriangulation2025}, but our previous work was limited to slowly moving target MAVs. It is important to study capturing highly maneuverable target MAVs. As the first work in this direction, we study how to achieve general aerobatic flight of capture MAVs. 

Aerobatic flight emphasizes leveraging the MAV's flexibility for complex, continuous maneuvers in confined spaces and requires high agility beyond speed optimization~\cite{kaufmannChampionlevelDroneRacing2023}. Although acrobatic flight control has been studied extensively~\cite{lupashinSimpleLearningStrategy2010, xieLearningAgileFlights2023, faesslerDifferentialFlatnessQuadrotor2018, sunComparativeStudyNonlinear2022}, existing methods face two critical limitations.
One limitation is that they are usually restricted to specific types of maneuvers and predefined flight trajectories. For example, in~\cite{talAerobaticTrajectoryGeneration2023}, it is necessary to arrange different maneuvers in a particular order and constrain the position and speed attributes in each maneuver. Then, a controller is used to track the desired trajectory.

Another limitation is the sim2real gap, which is not limited to aerobatic tasks but also occurs in various robot learning tasks. Specifically, the neural network trained in simulation performs poorly when deployed in real robots, hindering its transfer from simulation to real-world environments. 
For example, robots controlled by neural networks may perform well when moving in one direction but poorly in another~\cite{radosavovicRealworldHumanoidLocomotion2024}.

Motivated by the limitations of the existing methods, we propose a new target-and-command-oriented (TACO) reinforcement learning framework. 
In the TACO framework, a target status is used to extract specific invariant quantities in different maneuver tasks.
Besides, the task-command-oriented neural network controller enables flexible aerobatic maneuver adjustment.
Depending on the specific task, these elements have different meanings, enabling diverse maneuvering tasks to be trained within a uniform framework.

To solve the sim2real gap limitation, we first establish a high-fidelity MAV model that incorporates motor dynamics and aerodynamics in the simulation and then identify dynamic parameters in the real environment. 
Then, during training, spectral normalization is applied to the policy network to affect the gradient between input and output. 
By adjusting the key parameters, the network generates some special properties that allow us to implement zero-shot sim2real.

\begin{figure}
    \centering
    \includegraphics[width= 0.9\linewidth]{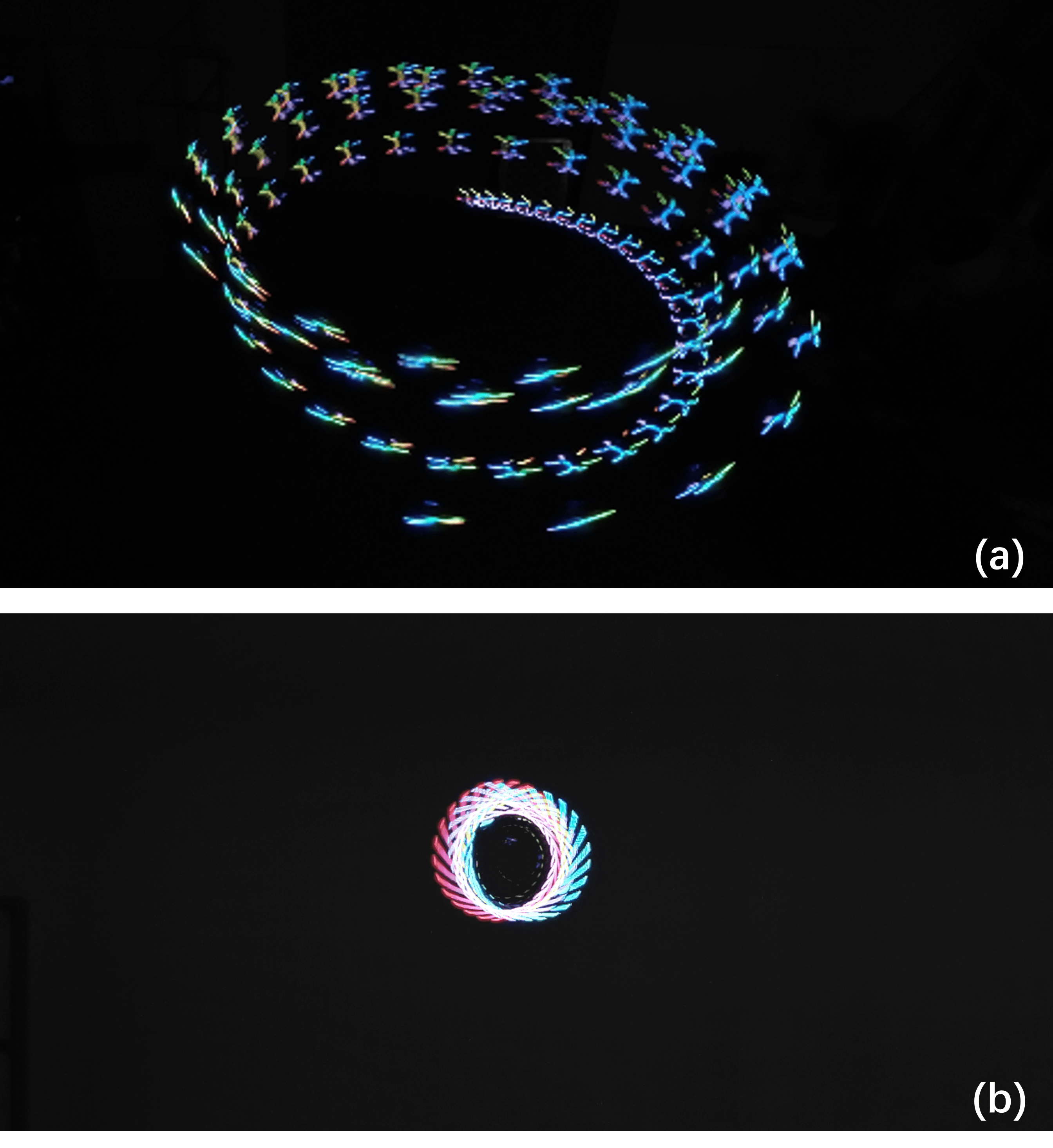}
    \caption{The real-world acrobatic flight trajectory based on TACO frame. (a) shows the flight trajectory of the CIRCLE task with the increasing desired speed. (b) shows the flight trajectory of the FLIP task with multi-flips.
    }\label{fig_whole_traj}
\end{figure}

\begin{figure*}
    \centering
    \includegraphics[width= \linewidth]{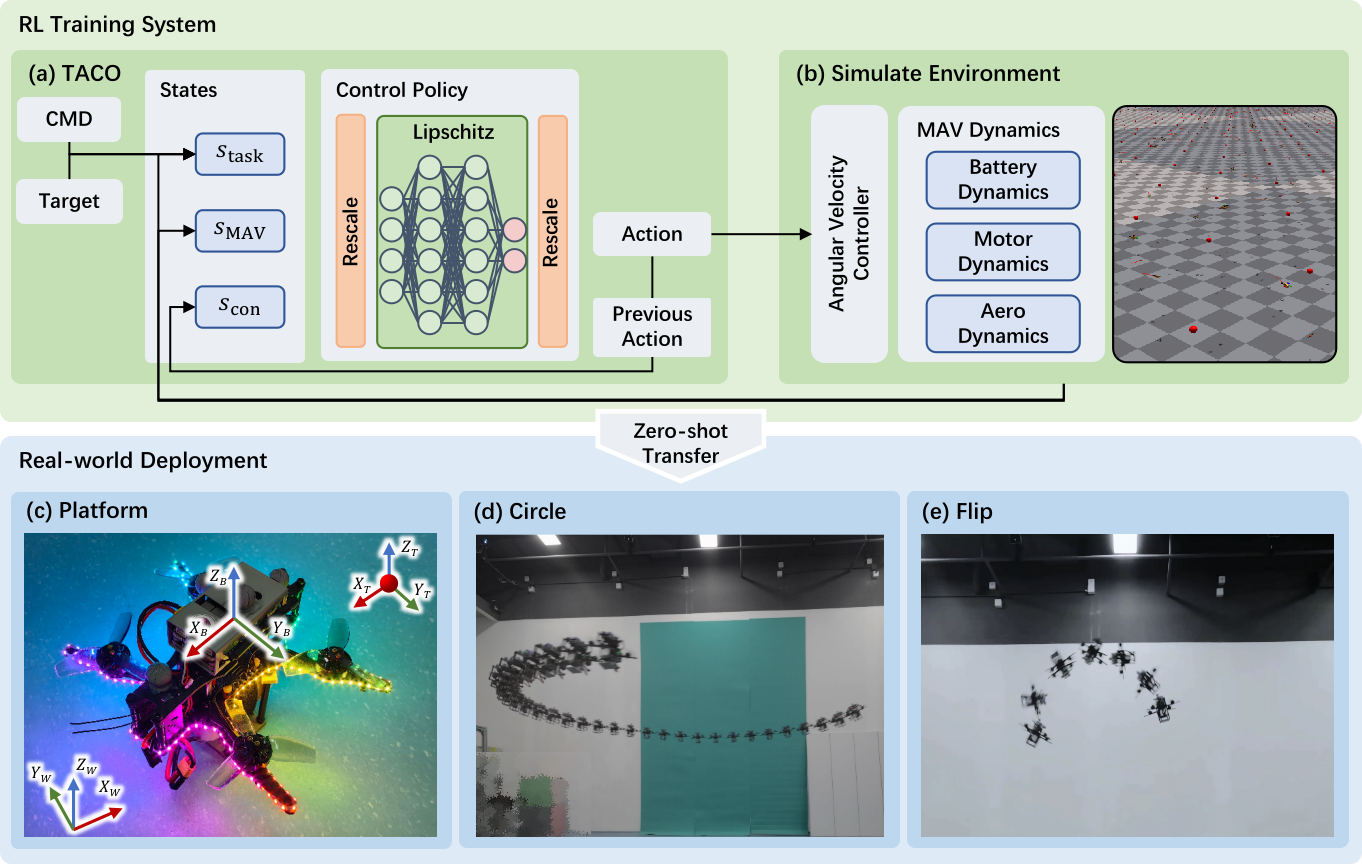}
    \caption{
    The overall structure of the TACO framework. 
    The higher section presents the RL training system, including the components of the TACO framework (a) and the simulation environment (b).
    (c) shows the relationships between the real MAV, the target status (ball in red), and the world frame.
    (d) and (e) show the results of the MAV executing CIRCLE and FLIP tasks in a real-world environment. 
    }
    \label{whole_sys}
\end{figure*}

Compared to the previous methods, the proposed TACO framework has the following novel features. 
First, it does not require predefined maneuver trajectories and supports adjusting flight parameters online.  Different aerobatic maneuvers can be learned with the unified form of state design.
Second, we introduce a novel method based on Lipschitz constraints and input-output rescaling. 
This method improves the policy's temporal and spatial smoothness, independence, and symmetry, thereby overcoming the sim2real gap in a zero-shot way without requiring complex dynamic models or reward functions.
These properties allow us to judge the network's performance without actually deploying it, saving time and preventing hardware damage.

Through real-world experiments, we demonstrate that TACO achieves superior performance in aggressive maneuvers, such as high-speed circular flights with a tilted attitude of more than 70 degrees and 14 stable fix-point continuous flips, with better tracking accuracy than traditional controllers such as model predictive controller (MPC).

\section{Related Work}\label{section_related_work}

\subsection{Aerobatic Flight for MAVs}
Many Aggressive maneuvers have been achieved like multi-flips~\cite{lupashinSimpleLearningStrategy2010}, flying through narrow gaps~\cite{falangaAggressiveQuadrotorFlight2017,xiaoFlyingNarrowGap2023,xieLearningAgileFlights2023}, perching on inverted surfaces~\cite{mellingerTrajectoryGenerationControl2012, falangaAggressiveQuadrotorFlight2017}, and combination of different maneuvers~\cite{kaufmannDeepDroneAcrobatics2020,talAerobaticTrajectoryGeneration2023}. 
Dividing the highly maneuverable fancy flight trajectory into small segments and tracking these segments one by one using the trajectory tracking controller is a common choice. In~\cite{kaufmannDeepDroneAcrobatics2020}, constrained polynomial trajectories are used for the MAV to enter, transition between, and exit the maneuvers. On the contrary, our controller learns the motion pattern instead of acting as a trajectory-tracking controller and does not require a dynamically feasible trajectory. 

For the Circle maneuver in this paper, prior studies achieved a radius of 1.8~m at 4~m/s~\cite{faesslerDifferentialFlatnessQuadrotor2018} and 4~m at 10~m/s~\cite{torrenteDataDrivenMPCQuadrotors2021}. 
Our experiments achieve a more aggressive maneuver with a 1.2~m radius at 5~m/s, yielding 4.2~rad/s angular velocity (1.6× faster than~\cite{torrenteDataDrivenMPCQuadrotors2021}) and 70 degrees tilt angle.

As for the FLIP maneuver, the 5-flip maneuver is first achieved in~\cite{lupashinSimpleLearningStrategy2010} through open-loop control methods that specify the thrust and desired angular velocities at each moment without relying on state feedback. In~\cite{chenLyapunovbasedControllerSynthesis2017, chenGenerationRealtimeImplementation2016, chenNonlinearAdaptiveControl2018}, single, double, and triple flips, which are also achieved in this paper, are achieved based on classic controllers, such as Lyapunov-stability based controller. 
To the best of our knowledge, reinforcement learning has not achieved a multi-flip maneuver. 
Besides, different from all the flip maneuvers in the above study, where the MAV first pulls up and then rotates and finally lowers the height and re-stabilizes, we achieve a stable fix-point continuous flip maneuver without losing altitude or pausing in the middle to re-stabilize.
The performance is more like a power loop maneuver with a tiny radius.

\subsection{Multi-Task Learning and Goal-oriented Learning}

Due to the ability to provide better flexibility and online adjustment, neural network controller with conditional inputs is used in legged robots~\cite{margolisWalkTheseWays2022} and drones~\cite{bauersfeldUserConditionedNeuralControl2023, xingMultiTaskReinforcementLearning2024} to achieve multi-task learning and goal-oriented learning.

In~\cite{bauersfeldUserConditionedNeuralControl2023}, researchers compared the effects of different network architectures on the performance of racing tasks with user-specified thrust-weight ratio and viewing direction. 
\cite{xingMultiTaskReinforcementLearning2024} is a work carried out at the same time as our work. 
They use complex networks and reward functions to learn to stabilize the quadrotor from high speed, autonomously race through a fixed track, and track randomly generated velocities.

Our framework also supports multi-tasking and goal-oriented learning.
Besides, we achieve better command tracking performance than MPC controllers with a unified design of state and reward functions and the only need for a simple, fully connected network. 

\subsection{Sim2Real for MAVs}

Some researchers have enhanced the fidelity of dynamic models to narrow the sim2real gap~\cite{torrenteDataDrivenMPCQuadrotors2021,songFlightmareFlexibleQuadrotor2021, bauersfeldNeuroBEMHybridAerodynamic2021}. 
However, these approaches require complex modeling and large datasets for parameter identification or neural network training, and the sim2real gap can never be eliminated~\cite{hoferPerspectivesSim2RealTransfer2020}.

Techniques like dynamics randomization and online identification have also been employed to improve policy's robustness and adaptability~\cite{songReachingLimitAutonomous2023,zhangLearningSingleHover2023} to handle the residual sim2real gap. 
Some researchers focus on increasing the robustness of controllers to environmental changes through designing state spaces, action spaces, and reward functions that are less sensitive to dynamic changes~\cite{kaufmannBenchmarkComparisonLearned2022, mysoreHowTrainYour2021,mysoreRegularizingActionPolicies2021}.
However, the performance of overcoming the sim2real gap can only be evaluated post-deployment, and the policy may not be perfectly symmetrical~\cite{radosavovicRealworldHumanoidLocomotion2024}. 

In contrast, our method directly optimizes intuitive and interpretable policy metrics, allowing for better prediction of deployment performance and better interpretability. 

\section{Methodology}
The overall system is illustrated in Fig.~\ref{whole_sys}. The system consists of three parts. The first part is the proposed TACO framework, and the second and third parts are the simulation environment and the actual MAV platform. In this section, we first introduce the dynamics of the MAV in the simulation environment, and then we introduce the MDP elements of the TACO framework and the training methods. The actual MAV platform is introduced in the experiment section.

\subsection{MAV Dynamics}
The MAV's status includes position $\p \in \mathbb{R}^3$, linear velocity $\vel \in \mathbb{R}^3$, attitude represented by a rotation matrix $\R \in \mathbb{R}^{3\times 3}$, angular velocity $\bomega \in \mathbb{R}^3$, motor speed $\Omega_i$ of motor $i$, battery voltage $V$. The dynamical equations are
\begin{align}
    \begin{matrix}
        \dot{\p} = \vel,  &  \dot{\vel} = \frac{\R^{-1}}{m} \left(\e_3 f_\Sigma  + \bk_\mathrm{drag} \vel^B\right) - \g, \\
        \dot \R = \R[\bomega^B]_\times,  & \dot{\bomega}^B = \mathbf{J} ^{-1}\left(\btau- [\bomega^B]_\times \mathbf{J} \bomega^B\right) ,  
    \end{matrix}
    \label{MAV_dynamics}
\end{align}
where $\g = [0, 0, 9.81]^T$ is the gravity acceleration, $m \in \mathbb{R}$ and $\mathbf{J} \in \mathbb{R}^{3 \times 3}$ are the mass and inertia, respectively, $\e_3 = [0,0,1]^T$ is a unit vector, and $\bk_\mathrm{drag} \in \mathbb{R}^{3 \times 3}$ is the drag coefficient matrix.
The superscript $B$ denotes the body frame of the MAV. 

The collective thrust $f_\Sigma$ and torque $\btau$ are calculated by
\begin{align}
f_\Sigma = \sum_{i=1}^4 f_{i}, \quad & \btau =\sum_{i=1}^4(\mathbf{r}_i \times \e_3 f_{i} + \tau_{i}), \\
f_{i} = k_\mathrm{force}\Omega_i^2, \quad & \tau_{i} = k_\mathrm{torque}\Omega_i^2 ,
\end{align}
where $f_{i}$ and $\tau_{i}$ are the thrust and torque generated by motor $i$, $\mathbf{r}_i$ is the arm of force. $k_\mathrm{force}$ and $k_\mathrm{torque}$ are aerodynamic parameters.

The motor is simulated using a first-order system. The dynamic equations with the motor label $i$ omitted are
\begin{align}
\dot\Omega &= \frac{1}{k_\mathrm{motor}}(\Omega_\mathrm{steady} - \Omega), \\
\Omega_\mathrm{steady} &= \mathrm{Poly}(V, \mathrm{PWM}), 
\end{align}
where $k_\mathrm{motor}$ is the motor constant, $\Omega_{\rm steady}$ and $\mathrm{PWM}$ are the steady speed and PWM (Pulse-Width Modulation) of each motor, respectively. 
$\mathrm{Poly}$ is a polynomial function obtained from the identification. 
Battery dynamics are referenced from~\cite{bauersfeldRangeEnduranceOptimal2022}.
The PWM on each motor is the output of the angular velocity flight controller, whose input is the collective thrust and the desired angular velocity.

\subsection{State Design}\label{section_state_design}
TACO introduces a unified state design method to handle different maneuvers discussed in this paper.
More specifically, the state $\s$ in different maneuvers under the TACO framework consists of three components:
\begin{align}
    \s = [\s_{\rm task}^T, \s_{\rm MAV}^T, \s_{\rm con}^T]\in\mathbb{R}^{26}.    
    \label{whole_state}
\end{align}
The three components are explained below.

\emph{\bf Task-oriented state} $\s_{\mathrm{task}}$ integrates all the information needed to describe and adjust a maneuvering task. 
It consists of two sub-states:
\begin{align} 
    \s_{\mathrm{task}} = [\s_{\rm r}^T, \s_{\rm t}^T] \in \mathbb{R}^{14}.
\end{align}
The first sub-state $\s_{\mathrm{r}}$ is given by $\s_{\mathrm{r}} = [\s_{\mathrm{r1}}^T, \s_{\mathrm{r2}}^T]\in\mathbb{R}^{12}$ where
\begin{align*}
    \s_{\mathrm{r1}} &= \p_{\mathrm{rel}}^B\in\mathbb{R}^{3},\quad   \s_{\mathrm{r2}} = {\rm vec}(\R_{\mathrm{rel}}^B)\in\mathbb{R}^{9},
\end{align*}
where ${\rm vec}(\cdot)$ converts a matrix into a vector.

Here, $\p_{\mathrm{rel}}^B = {\R}^{-1} (\p_{\rm v} - \p)$ and $\R_{\mathrm{rel}}^B = {\R}^{-1} \R_{\rm v}$ respectively represent the relative position and relative orientation between the MAV' state and a \emph{\bf target state} that consists of $\p_{\rm v}$ and $\R_{\rm v}$. 

The target state is an important concept in the TACO framework. 
It ensures the state representation's consistency and reduces the reward function design's difficulty. 
Details on designing the target state for each task will be provided shortly.

It is worth noting that we use the rotation matrix to represent the attitude instead of Euler angles or quaternions. That is because research shows that rotation matrices are more suitable for deep learning applications~\cite{zhouContinuityRotationRepresentations2019}. 
We further verify the superiority in the experiment section.

The second sub-state $\s_{\rm t}$ is $\s_{\rm t} = [\s_{\rm t1}, \s_{\rm t2}]\in\mathbb{R}^{2}$, where $\s_{\rm t1}\in\mathbb{N}$ is the task flag that specifies the type of maneuver, and $\s_{\rm t2}\in\mathbb{R}$ specifies the command of the task, which is used for adjusting maneuver parameters during task execution.

To illustrate the design of the above task-oriented state, we consider the hover task (referred to as POS) first and then two types of maneuvers: CIRCLE and FLIP. 

1) In the POS task, the MAV should fly to and hover at a desired position $\p^*$ with the desired yaw $\psi^*$.

The task flag is set for the POS task to $\s_{\rm t1} = 0$. The target state is set as $\p_{\rm v} = \p^*$ and $\R_{\rm v}=\mathcal{R}(0, 0, \psi^*)$. 
The target's roll and pitch are set to 0 to ensure feasible hovering.
Since the POS task does not require additional commands, we can set $\s_{\rm t2} = 0$ as a mask to maintain consistency with other tasks.

2) In the CIRCLE task, the MAV should rotate in a horizontal plane around the center point at $\p^*$ with the desired speed $v^*$ and desired radius $r^*$ while heading to the center point. 

The task flag is set to $\s_{\rm t1} = 1$ for the CIRCLE task task. The target state is selected as $\p_{\rm v} = \p^*$ and $\R_{\rm v}=\mathcal{R}(0, 0, 0)$. 
To make the desired tangential speed $v^*$ adjustable, we set $\s_{\rm t2} = v^*$. Although technically feasible, we ignore the adjustability of the radius.

3) In the FLIP task, the MAV should first hover at the desired position $\p^*$ with the desired attitude $\R^* = \mathcal{R}(0, 0, 0)$.
It then performs $k$-flips around its x-axis to achieve a total rotation angle of $2k\pi$.
Finally, the MAV returns to the hover state.

For the FLIP task, the task flag is set to $\s_{\rm t1} = -1$. 
The target state is selected as $\p_{\rm v} = \p^*$, $\R_{\rm v} = \R^*$. 
To measure continuous flips, we use the cumulative change in the roll angle as the rotated angle. Each time the drone flips, the rotated angle increases or decreases by $2\pi$.
We take the remaining angle to be rotated $\Delta \phi$ as the task command, namely, $\s_{\rm t2} = \Delta \phi$.

\emph{\bf MAV-oriented state} $\s_{\rm MAV}$ is used to describe the MAV's motion and safety information and is defined as
\begin{align}
    \s_{\rm MAV} = [(\vel^B)^T, (\bomega^B)^T, h, V]^T \in \mathbb{R}^{8},
\end{align}
where $h$ represents the MAV's altitude used for safety concerns. 
Battery voltage $V$ detects power consumption and voltage drops under extreme maneuvering.

\emph{\bf Context-oriented state} $\s_{\rm con}$ represents the previous action:
\begin{align} 
\s_{\mathrm{con}} = \a_{t-1} \in \mathbb{R}^{4}.
\end{align} 
Action is selected as the collective throttle and body rate $\a = [f_\Sigma, \boldsymbol{\omega}^T] \in \mathbb{R}^{4}$. 
The collective throttle is an abstract quantity ranging from 0 to 1000.
$\s_{\mathrm{con}}$ is to account for delays in sensor sampling, model inference, and signal transmission.

\subsection{Reward Design}

In this subsection, we design the reward functions for the POS, CIRCLE, and FLIP tasks in a unified manner. 
The total reward function is the product of sub-reward functions. 
A sub-reward function has the following unified expression:
\begin{align}
r(x, \lambda, n) = \sum_{i=1}^{n} \frac{1}{1 + \lambda_i {||x||}^2},    
\label{frac_reward}
\end{align}
where $x$ is the quantity expected to be minimized and designed for a given maneuver task, $\lambda_i \in \mathbb{R}$ and $n \in \mathbb{N}$ are coefficients that should be turned. 
An example will be given later to demonstrate.
For simplicity, $r(x, \lambda, n)$ is denoted as $r(x)$. 

1) The reward function for the POS task is designed as
\begin{equation}
    \begin{aligned}
        r_{\rm p} = r(\p_{\mathrm{rel}}^B)r(\alpha) ,
    \end{aligned}
    \label{pos_reward}
\end{equation}
where $\alpha$ is the angle from the axis–angle representation of the relative attitude of the body and target frames.

2) The reward function for the CIRCLE task is
\begin{equation}
    \begin{aligned}
        r_{\rm c} = r(e_3^T\p_{\mathrm{rel}})r(||\p_{\mathrm{rel}} - e_3^T\p_{\mathrm{rel}}|| - r^*)r(v_\bot - v^*)r(\alpha) ,
    \end{aligned}
    \label{circle_reward}
\end{equation}
where $v_\bot$ represents the component of $\vel$ that is perpendicular to $\p_{\mathrm{rel}}$ in the XOY plane, $\alpha$ is angle between the x axis of body frame and $\p_{\mathrm{rel}}$. 

3) The reward function for the FLIP task is
\begin{equation}
    \begin{aligned}
        r_{\rm f} = r(\p_{\mathrm{rel}}^B)r(\alpha)r(\Delta \phi) ,
    \end{aligned}
    \label{flip_reward}
\end{equation}
where $\alpha$ represents the angular deviation between the X-axes of the body frame and the target frame.

\begin{figure*}
    \centering
    \includegraphics[width= \linewidth]{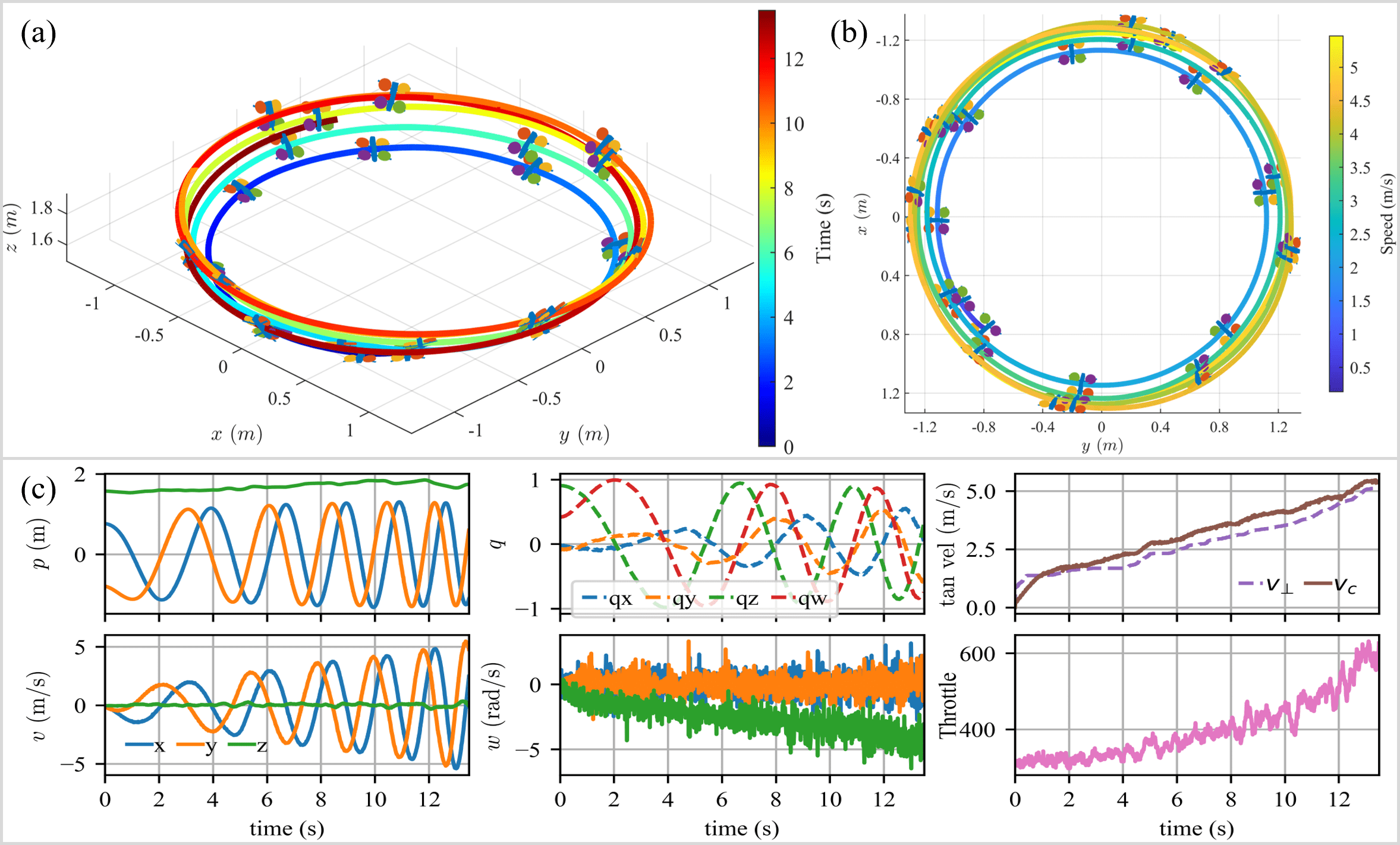}
    \caption{Real-world flight trajectories under different viewing angles and flight state curves in the CIRCLE task. 
    (a) shows the 3D flight trajectory with the color indicating the time step. (b) shows the XOY 2D flight trajectory with the color indicating the speed. (c) shows the MAV's state and commands in order.
    }
    \label{circle_traj}
\end{figure*}

\subsection{Policy Training Method}\label{section_policy_design}

We use a $L$-layer fully connected network as the policy, where the last layer uses the Tanh activation function while the remaining layers use the ReLU activation function.
The state $\s$ is rescaled by $\bk_i \in \mathbb{R}^{26}$ before being sent into the network, and the output of the network is rescaled by $\bk_o \in \mathbb{R}^{4}$ to get $\a$. 
We apply the Lipschitz constant $k_{\rm Lip}$ to all layers.

Considering the entire network as a composite function $h$ mapping from $\s$ to $\a$, since ReLU and Tanh satisfy the 1-Lipschitz constraint, the scaling relationship satisfies the following inequality:
\begin{equation}
    \begin{aligned}
        \frac{||h(\s_1) - h(\s_2)||}{||\s_1 - \s_2||} \leq k_{\rm Lip}^L||\bk_i||||\bk_o||, 
    \end{aligned}
    \label{lip_constrain}
\end{equation}
where $\s_1$ and $\s_2$ are two arbitrary different states. 

The gradient from $\s$ to $\a$ is also affected by $k_{\rm Lip}$ and scaling parameter $\bk_i,\bk_o$. 
With suitable parameters, the policy can exhibit the following properties.
\begin{itemize}
    \item[1)] \emph{\bf Temporal Smoothness}: The action sequence is continuous over time.
    \item[2)] \emph{\bf Spatial Smoothness}: When the state changes smoothly, the action also changes smoothly.
    \item[3)] \emph{\bf Independence}: unrelated actions remain unchanged during task execution.
    \item[4)] \emph{\bf Symmetry}: When the states are symmetric, the actions are also symmetric.
\end{itemize}

These properties can greatly help overcome the sim-to-real gap when it can not be further narrowed with system identification, as demonstrated in our experiments.
Specifically, after each parameter update, we calculate the maximum singular value of the parameter matrix of each policy layer. 
If it is greater than the Lipschitz constraint, all parameters are scaled down by the Lipschitz constraint.

\begin{figure*}
    \centering
    \includegraphics[width= \linewidth]{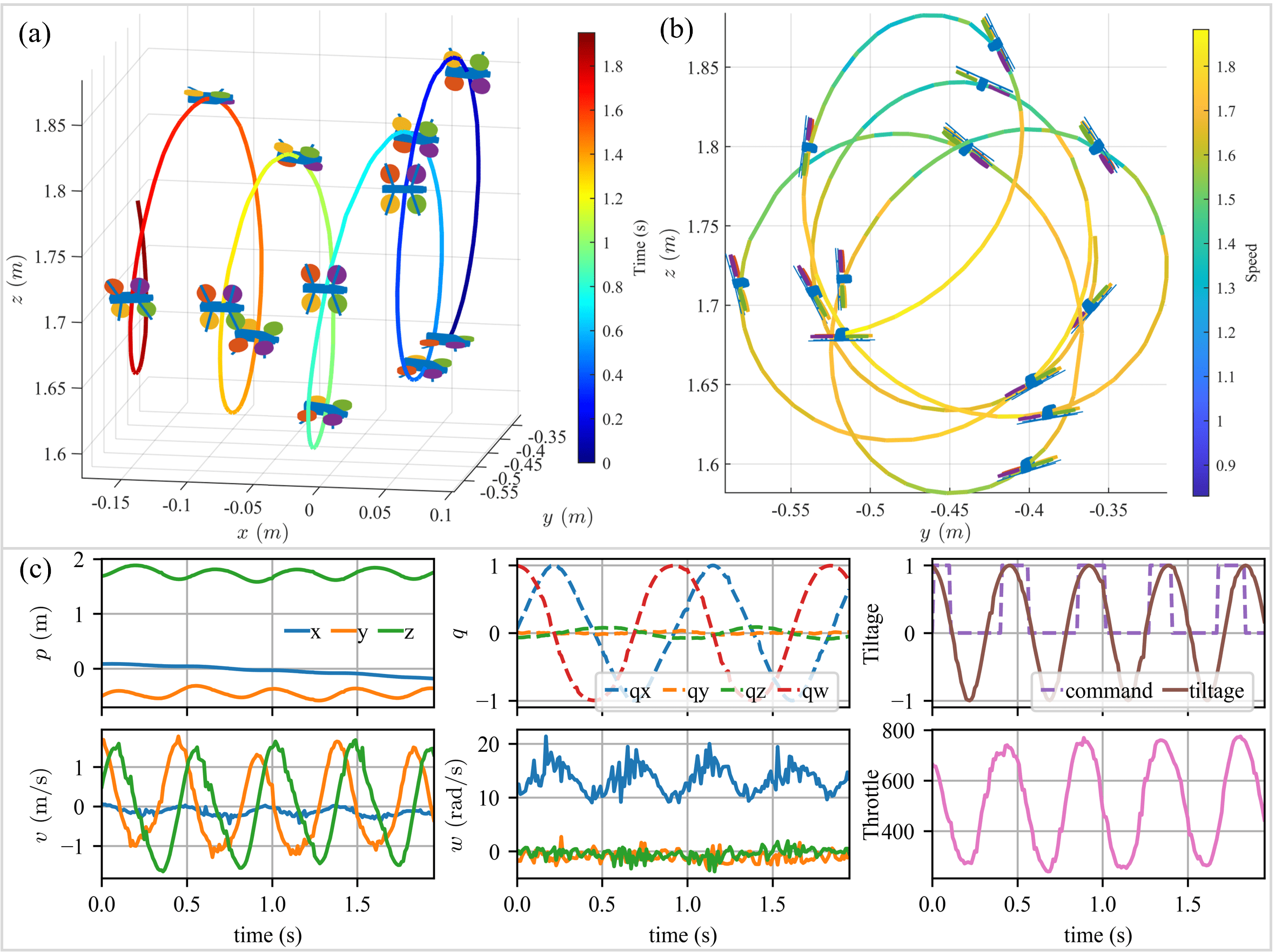}
    \caption{Real-world flight trajectories in the real world under different viewing angles and flight state curves in the FLIP task. (a) shows the 3D flight trajectory with the color indicating the time step. (b) shows the YOZ 2D flight trajectory with the color indicating the speed. (c) shows the MAV's state and commands in order.
    }
    \label{flip_traj}
\end{figure*}

\section{Experiments}
In this section, we validate the effectiveness of the TACO framework through a series of experiments.
First, we describe the experiment setup.     
Next, we evaluate the policy's performance in the CIRCLE and FLIP tasks on real MAVs.    
Then, we evaluate how our training method influences the policy's properties with the POS task.
Finally, we compare the command tracking performance of the TACO framework and the MPC on the CIRCLE task.
The differences between our approach and related work based on other classical control theories and the progress we have made on various indicators for quantitative comparison are also described in section~\ref{section_related_work}. 

\subsection{Experiments Setup}
The MAV in Fig.~\ref{whole_sys} (c) has a mass of 0.46~kg, a motor-to-motor distance of 0.149~m, and a thrust-to-weight ratio 4.1. 
Onboard sensors provide voltage data at 1000~Hz, while a Vicon system captures real-time position, attitude, and velocity information at 200~Hz. 
The policy inferences at 100~Hz, and the angular velocity flight controller is executed at 1000~Hz.

The policy is trained based on 2048 parallel environments in IsaacGym simulator~\cite{makoviychukIsaacGymHigh} augmented with dynamics introduced before. 
Due to space limitations, dynamic parameters identified through experiments and training parameters are omitted.
During training, the actor receives noisy data while the critic receives the ground truth.
When an episode starts, the MAV's initial position, orientation, velocity, motor speeds, task commands, and dynamic parameters are randomized for each environment. 
The randomization range expands with the training progress, implementing a simple yet effective curriculum learning.
The episode ends when the altitude of the MAV is under the threshold or the time limitation is reached.

\subsection{Effectiveness of Target-And-Command-Oriented State}

This section uses the CIRCLE and FLIP tasks in the real world to verify the TACO framework's ability to adjust online task commands.

\subsubsection{CIRCLE task}

We conduct the CIRCLE task on a real MAV platform, where the desired radius is set to $1.2$~m, and the desired tangential velocity $v_{\rm c}$ varies within the range of $[-5, 5]$~m/s. 

For the analysis, we record experimental data where $v_{\rm c}$ increases from $0$~m/s to $5$~m/s in $14$~s. 
Fig.~\ref{circle_traj}~(a) and (b) present the flight trajectories from different perspectives, while Fig.~\ref{circle_traj}~(c) illustrates the status of the MAV over time.


Regarding position, the MAV follows a circular trajectory, with the radius and center aligning well with the desired path.
Regarding attitude, the MAV consistently maintains its orientation toward the target point. 
The tilt angle gradually increases as the speed increases, reaching more than $70$~degrees. 
For angular velocities, the roll and pitch rates remain close to $0$, while the yaw rate increases to $5$~rad/s, though accompanied by slight oscillations.
Finally, in terms of average throttle, the throttle increases from $300$ to $600$, which corresponded well with the actual velocity trend.



\subsubsection{FLIP task}



Next, we test the FLIP task on the same MAV platform. 
To push the controller to its limits, we issue continuous flip commands. 
The video shows a continuous 14-flips finished within $6.6$~s.
As the number of flips increases, the MAV's movements become more stable. 
Because of its stable flight, the MAV can flip until its battery runs out.
Based on the battery model and real-world experiment data, about 90 flips can be executed.

Fig.~\ref{flip_traj}~(a) and (b) show the flight trajectories of 4 flips out of the 14-flips from different perspectives, Fig.~\ref{flip_traj}~(c) shows the changes in both the task commands and the MAV's status.

Regarding task commands, the MAV must flip once each time the command jumps from $0$ to $1$. 
The commands are triggered by human operators randomly without specific frequency.
The "tiltage" metric represents the MAV's attitude, with a value of $1$ indicating right side up and $-1$ indicating completely inverted. 

Regarding position variation, the MAV maintains stability in the YOZ plane, with displacement perpendicular to this plane remaining within approximately 5~cm, demonstrating the controller's ability to restrict vertical deviations effectively.
Regarding attitude, the X-axis of the MAV is basically the same direction as the X-axis of the world frame, indicating that the MAV maintains precise control during the flips. 
Regarding angular velocities, the roll and yaw rates remain near zero, while the pitch rate periodically fluctuates between $10$ and $20$ rad/s, corresponding to the continuous flips.

\subsection{Sim2Real Performance Evaluation}
We assess the policy's spatial smoothness, independence, and symmetry in simulation and evaluate temporal smoothness through real-world experiments.

\begin{figure}
    \centering
    \includegraphics[width=\linewidth]{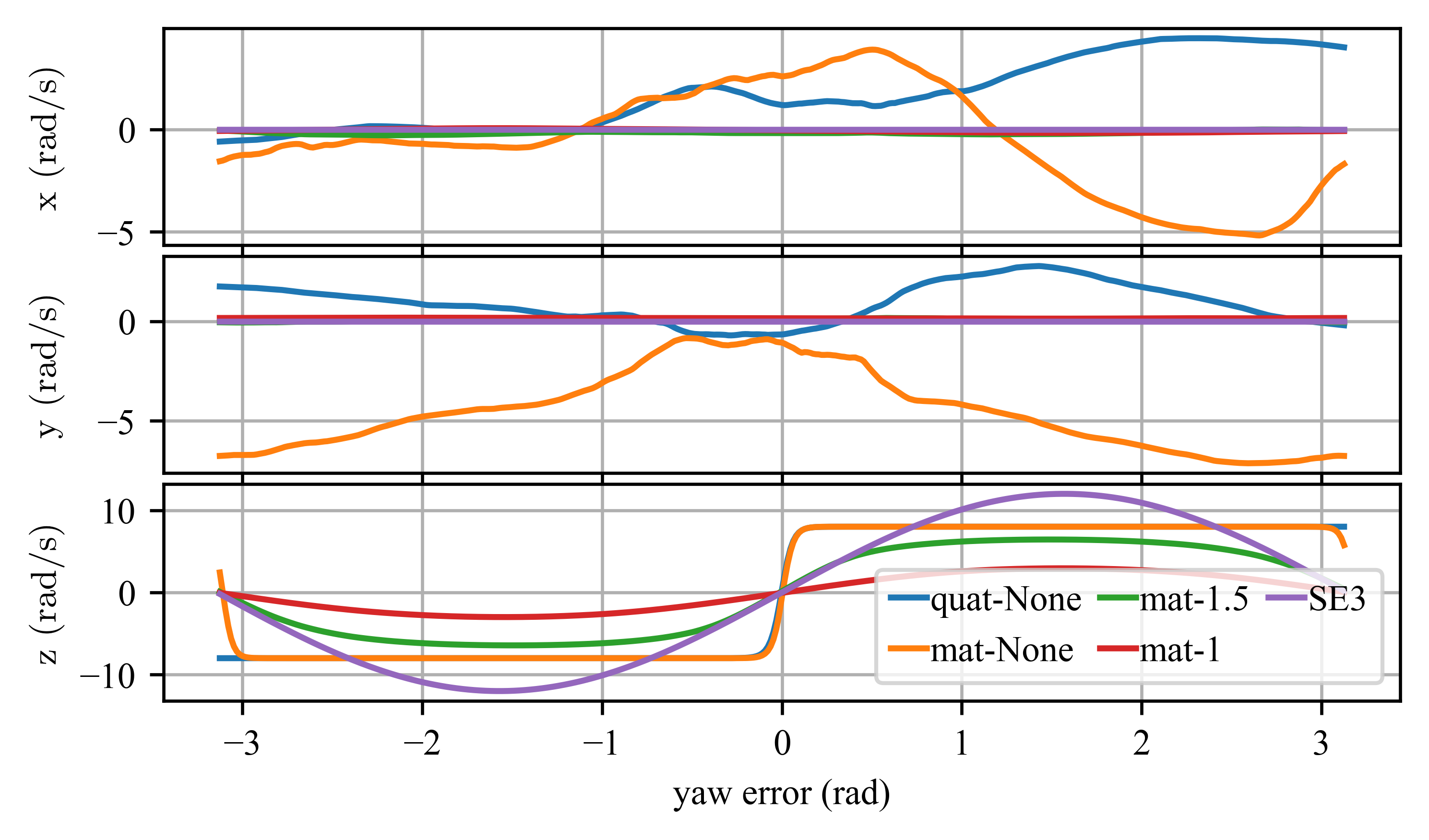}
    \caption{Desired angular velocity output by different policies with respect to the YAW deviation in $(-\pi,\pi)$~rad. 
    Label 'quat' represents a policy using quaternions, and 'mat' represents a policy using a rotation matrix. 
    'None' indicates that the Lipschitz constraint is not used, and '1' and '1.5' indicate the Lipschitz constant.}
    \label{different_mat_1}
\end{figure}

\begin{figure}
    \centering
    \includegraphics[width=\linewidth]{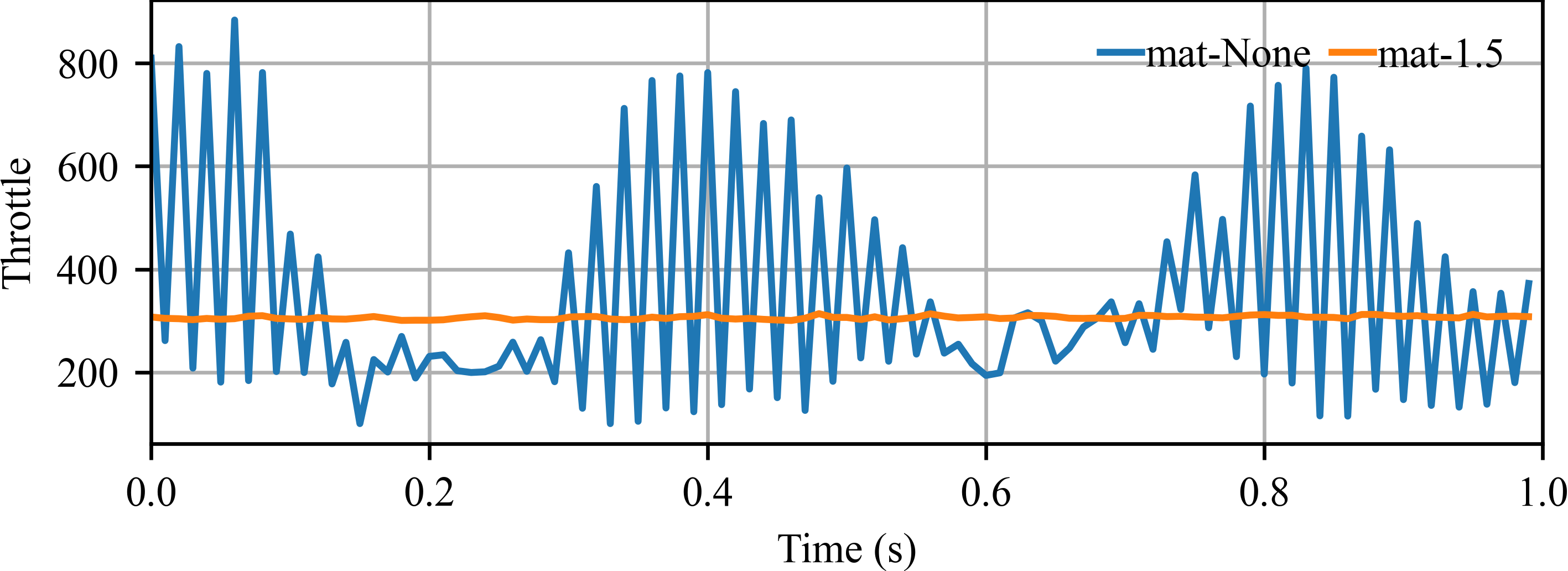}
    \caption{Average throttle sequence of policy "mat-None" and "mat-1.5".}
    \label{different_mat_2}
\end{figure}

\subsubsection{Evaluation in simulation}
We train policies based on quaternion and rotation matrix representations with different Lipschitz constants. 
Additionally, we use the SE3 controller as a comparison. 

After training, we place the MAV in the desired position and change its attitude so that the yaw error between the MAV's attitude and the desired attitude changes at $(-\pi,\pi)$. 
The curves of the desired angular velocity with respect to yaw error for different controllers are shown in Fig.~\ref{different_mat_1}.

Regarding spatial smoothness, as the yaw error increases, the desired angular velocity output by the "quat-None" policy for the z-axis quickly reaches its boundary value. 
Although yaw errors of $+\pi$ and $-\pi$ correspond to nearly identical attitudes, the desired angular velocities are opposite, indicating a lack of spatial smoothness. 
In contrast, the "mat-None" policy outputs a desired angular velocity that first increases and then decreases, resulting in similar values at yaw errors of $+\pi$ and $-\pi$, thus demonstrating better spatial smoothness. 
When the Lipschitz constraint is introduced during training, the policy outputs a z-axis angular velocity of $0$ at both $+\pi$ and $-\pi$ yaw errors, further enhancing spatial smoothness.

With the Lipschitz constraint applied, the controller no longer produces extreme angular velocities. 
As the Lipschitz constant decreases, the maximum angular velocity also reduces.

Regarding independence, while only z-axis angular velocity is needed to eliminate yaw error, only policies with the Lipschitz constraint generate zero angular velocities along the x and y axes. 
Therefore, introducing the Lipschitz constraint improves the policy's independence.

For symmetry, the policy should produce opposite desired angular velocities for opposite yaw errors. 
A comparison between policies with and without the Lipschitz constraint shows that the Lipschitz constraint can significantly enhance symmetry.

Moreover, the SE3 controller performs well in spatial smoothness, independence, and symmetry. 
The output patterns of the policy based on the rotation matrix and Lipschitz constraint closely resemble those of the SE3 controller, suggesting that the agent has learned a similar control method, thereby enhancing the network's interpretability.

\subsubsection{Evaluation in deployment}

We deploy "mat-None" and "mat-$1.5$" in Fig.~\ref{different_mat_1} to a real MAV and record the throttle series shown in Fig.~\ref{different_mat_2}. 
It can be observed that, for targets at the same position, the throttle sequence generated by the "mat-None" policy exhibits more fluctuations, leading to poorer temporal smoothness performance compared to "mat-$1.5$."

\subsection{Command Tracking Performance}
Finally, we compared the TACO framework with the MPC on the CIRCLE task. 
The linear MPC calculates the acceleration based on the trajectory, and the SO3 controller calculates the thrust and torque based on the acceleration.
We set up 9 CIRCLE tasks with different fixed tangential velocity commands. 
After the drone's flight becomes stable, we record the actual radius and tangential velocity over 20~s and compare it with the expected value. 
The Mean Squared Error (MSE) of radius and velocity are shown in Table~\ref{speed tracking error}.


\begin{table}[ht]
\caption{MSE of radius and velocity of TACO framework and MPC under different desired velocity}
\centering
\begin{tabular}{ccc cc}
\toprule
\multirow{2}{*}{\textbf{Desired Velocity (m/s)}} & \multicolumn{2}{c}{\textbf{Radius MSE ($\rm m^2$)}} & \multicolumn{2}{c}{\textbf{Velocity MSE($\rm \frac{m^2}{s^2}$)}} \\
\cmidrule{2-5}
 & \textbf{TACO} & \textbf{MPC} & \textbf{TACO} & \textbf{MPC} \\
\midrule
\textbf{-4} & 0.0148 & \textbf{0.0076} & \textbf{0.1176} & 0.6737  \\
\textbf{-3} & \textbf{0.0019} & 0.0030 & \textbf{0.0070} & 0.2893 \\
\textbf{-2} & \textbf{0.0003} & 0.0100 & \textbf{0.0125} & 0.0816 \\
\textbf{-1} & \textbf{0.0036} & 0.0219 & \textbf{0.0097} & 0.1052 \\
\textbf{1} & \textbf{0.0007} & 0.0021 & \textbf{0.0439} & 0.0519 \\
\textbf{2} & \textbf{0.0014} & 0.0262 & \textbf{0.0843} & 0.1668 \\
\textbf{3} & \textbf{0.0015} & 0.0329 & \textbf{0.0087} & 0.3547 \\
\textbf{4} & \textbf{0.0168} & 0.0333 & \textbf{0.0199} & 0.6846 \\
\bottomrule
\end{tabular}
\label{speed tracking error}
\end{table}

As the table shows, TACO and MPC increase their radius and velocity tracking errors as the desired speed increases. 
However, the taco framework achieved a much smaller radius tracking error in most experiments and a more minor velocity tracking error in all experiments.
As a result, the TACO framework has better command tracking performance.

In addition, we note that when switching from hover to CIRCLE tasks, if the trajectory does not contain entry and exit parts, MPC-based drones are more likely to accelerate suddenly, resulting in a crash. 
In contrast, TACO-controlled drones have a smoother motion.
We believe this is because, in reinforcement learning training, the randomization of the UAV's initial state optimizes the performance at the beginning of the maneuver.

\section{CONCLUSIONS}

In this paper, we present the TACO framework, a novel approach designed to enhance the acrobatic flight capabilities of MAVs for aggressive aerobatic maneuvers. 
It supports real-time maneuver adjustments, offering greater flexibility in dynamic environments. 
Extensive simulations and real-world experiments, including high-speed, high-accuracy circular flights and stable fix-point continuous flips, demonstrate the impressive performance of our framework.
Future work will focus on expanding the framework's applicability to a broader range of aerial tasks and environments, further enhancing its robustness and generalization capabilities.


\bibliographystyle{ieeetr}
\bibliography{main}

\vfill

\end{document}